\documentclass[pdflatex]{article}
\usepackage{spconf,amsmath}
\usepackage{graphicx}
\usepackage{bm}
\usepackage{url}
\usepackage{amssymb}
\usepackage{booktabs}
\usepackage{cite}
\usepackage{subfigure}

\usepackage{float}
\usepackage[ruled, vlined]{algorithm2e}
\usepackage{algpseudocode}

\usepackage{multirow}

\newcommand{\ldotfill}[2]{\leavevmode\xleaders\hbox{\rule{2pt}{0.4pt}\ }\hfill\null}
\title{GHFP: Gradually Hard Filter Pruning}

\name{Linhang Cai${}^{\star}$${}^{\dagger}$, Zhulin An${}^{\star}$, Yongjun Xu${}^{\star}$}
\address{
${}^{\star}$Institute of Computing Technology, Chinese Academy of Sciences, Beijing, China\\
${}^{\dagger}$University of Chinese Academy of Sciences, Beijing, China\\
}

  \def\Hline{
  \noalign{\ifnum0=`}\fi\hrule \@height 4.\arrayrulewidth \futurelet
   \reserved@a\@xhline}

\begin{document}
\ninept
\maketitle

\begin{abstract}
%The abstract should contain about 100 to 150 words.
Filter pruning is widely used to reduce the computation of deep learning, enabling the deployment of Deep Neural Networks (DNNs) in resource-limited devices. Conventional Hard Filter Pruning (HFP) method zeroizes pruned filters and stops updating them, thus reducing the search space %and the capacity 
of the model. On the contrary, Soft Filter Pruning (SFP) simply zeroizes pruned filters, keeping updating them in the following training epochs, thus maintaining the capacity of the network. However, SFP, together with its variants, % SofteR Filter Pruning (SRFP), Asymptotic Soft Filter Pruning (ASFP) and Asymptotic SofteR Filter Pruning (ASRFP), 
converges much slower than HFP %in case of large pruning rates 
due to its larger search space. Our question is whether SFP-based methods and HFP can be combined to achieve better performance and speed up convergence. Firstly, we generalize SFP-based methods and HFP to analyze their characteristics.
Then we propose a Gradually Hard Filter Pruning (GHFP) method to smoothly switch from SFP-based methods to HFP during training and pruning, thus maintaining a large search space at first, gradually reducing the capacity of the model to ensure a moderate convergence speed. Experimental results on CIFAR-10/100 show that our method achieves the state-of-the-art performance.% and also outperform conventional methods in terms of subjective quality.
\end{abstract}


\vspace{-0pt}

%%%%%%%%%%%%%%%%%%%%%%%%%%%%%%%%%%%%%%%%%%%%%%%%%%%%%%%%%%%%%%%%%%%%%%%%%%%%%%%%%%%%%%%%%%%%%%%%%%%%%%%%%%%%%%%%%%%%%%%%%%%%%%%%%%%%%%%%%%%%%%%%
\section{Introduction}
\label{sec:intro}

Despite the success of deep Convolutional Neural Networks (CNNs) in visual tasks from image classification~\cite{He2016, HCGNet} to object detection~\cite{Girshick2014}, their expensive computation cost and memory footprint hinder their deployment in resource-limited devices. Thus, it is necessary to compress DNN models with little loss in performance. 

A typical filter pruning pipeline is composed of four phases: training a network, evaluating the importance of every filter, pruning unimportant filters and fine-tuning. The fine-tuning phase is to compensate for the performance loss caused by pruning. Prevalent filter pruning methods can be divided into two categories: hard filter pruning (HFP) and soft filter pruning (SFP)~\cite{He2018}. While SFP zeroizes pruned filters and updates them in the following training epochs to maintaining the capacity of the network,  HFP would not update those pruned filters any more. 

A disadvantage of SFP is that there will be a severe accuracy drop after pruning when the pruning rate is large due to its large search space. There are three variants of SFP to alleviate this issue, named SofteR Filter Pruning (SRFP)~\cite{Cai2020}, Asymptotic Soft Filter Pruning (ASFP)~\cite{He2019} and Asymptotic SofteR Filter Pruning (ASRFP)~\cite{Cai2020} respectively. ASFP gradually increases the pruning rate from zero to the final pruning rate to alleviate the accuracy drop caused by pruning, while SRFP decays
the pruned weights with a monotonic decreasing parameter to soften the effect of pruning. ASRFP is a combination of ASFP and SRFP to simultaneously increase the pruning rate and use a decreasing parameter. For clarity, we use ASP to denote ASFP and ASRFP.

\begin{figure}[t]
	\center
	\vspace{-4mm}
	\subfigure[Pruning Rate = 0.2]{
		\label{fig:soft_and_hard_comp_08}
		\includegraphics[width=0.5\linewidth]{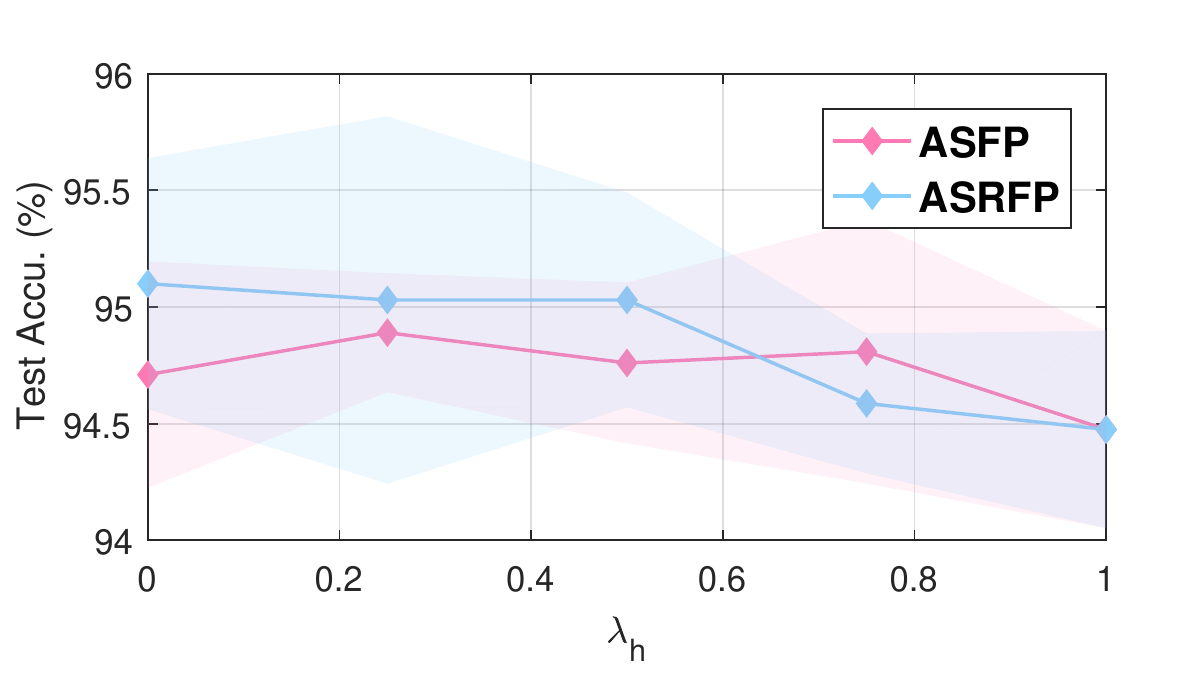}
	}
	\hspace{-5mm}
	\subfigure[Pruning Rate = 0.8]{
		\label{fig:soft_and_hard_comp_02}
		\includegraphics[width=0.5\linewidth]{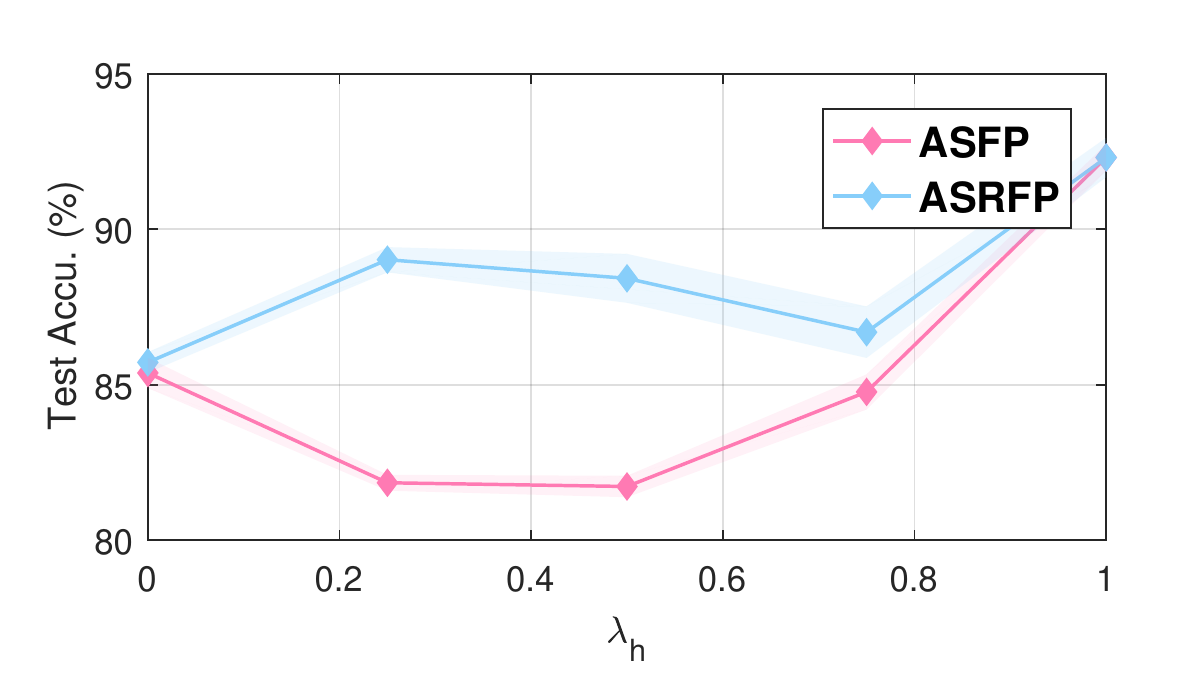}
	}
	\vspace{-2mm}
	\caption{
		Comparison of CIFAR-10 test accuracies of different $\lambda_{h}$ and pruning rates for ResNet-110, where $\lambda_{h}$ is the ratio of HFP to the total pruning rate. $\lambda_{h} =0$ means pure ASP, while $\lambda_{h} =1$ means pure HFP. (Solid line and shadow represent the mean and standard deviation respectively.)
	}
	\vspace{-1mm}
	\label{fig:soft_and_hard_comp}
\end{figure}

However, these soft pruning methods could not avoid this severe accuracy drop because they maintain a large search space, while HFP would gradually reduce the capacity of the model to ensure convergence. As shown in Figure\,\ref{fig:soft_and_hard_comp}, while for small pruning rates like 0.2, ASP may outperform HFP, for large pruning rates like 0.8, HFP is more advantageous. Specifically, for small pruning rates, the overall trend of the test accuracy is downward as the ratio of HFP increases from 0 to 1, meaning that ASP may be superior to HFP for small pruning rates, while the overall trend is upward for large pruning rates.

\begin{figure*}[ttt]
  \centering
  \includegraphics[width=90mm,clip]{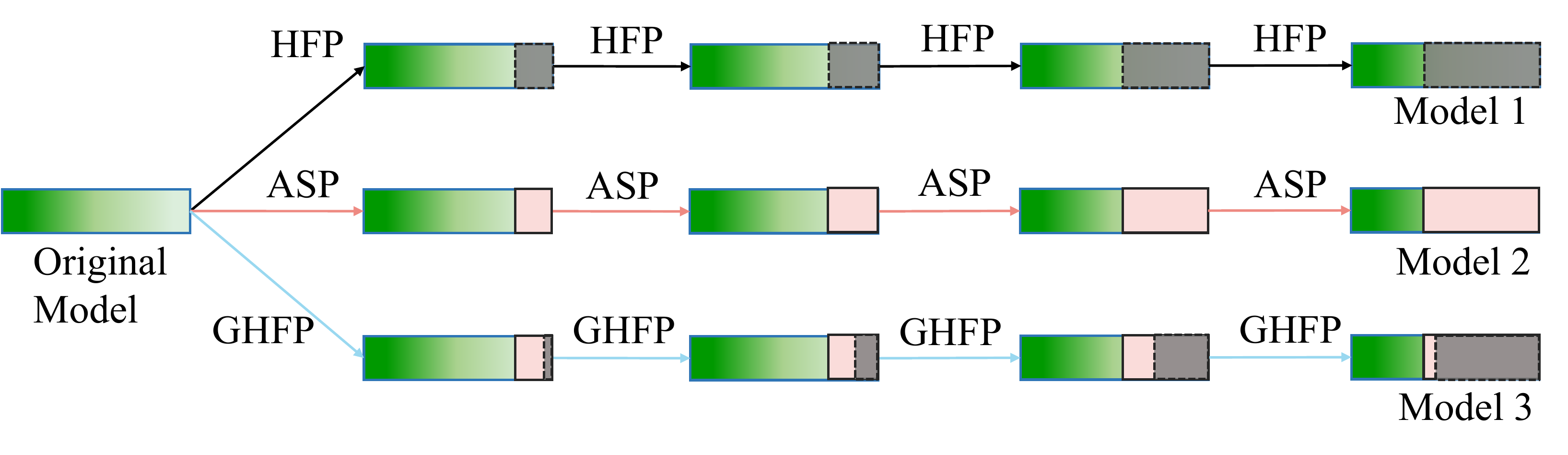}  
  \vspace{-5pt}
  \caption{Overview of our GHFP method. We use "ASP" to denote ASFP and ASRFP as they belong to the general soft filter pruning and they gradually increase the pruning rate from zero to the final pruning rate as HFP does. Green, gray and pink rectangles represent unpruned filters, hard pruned filters and softly pruned filters respectively. Deeper Green means more important filters. Softly pruned filters are allowed to update their parameters in the following training epochs, while hard pruned filters are not. Thus, ASP maintains a large search space, having trouble in obtaining a really compact model as HFP, which gradually reduces the search space to ensure convergence. Our GHFP uses a monotonic increasing parameter to control the ratio of HFP, increasing from 0 to 1 to balance performance and convergence speed. }
  \label{fig:ghfp}
  \vspace{-2pt}
\end{figure*}

Thus, we propose a Gradually Hard Filter Pruning (GHFP) method to smoothly switch from ASP to HFP during training and pruning to achieve a balance between performance and convergence speed, as shown in Figure\,\ref{fig:ghfp}. We utilize a monotonic increasing parameter to control the ratio of HFP, increasing from 0 to 1, thus acting like ASP to maintain a large capacity at start, gradually increasing the rate of HFP to reduce the search space to ensure convergence. 

Our contributions are as follows: (1) We generalize ASP and HFP to analyze their characteristics, noting that ASP maintains a large search space at the cost of much slower convergence speed than that of HFP. (2) We propose GHFP to smoothly switch from ASP to HFP during training and pruning, performing well across various networks, datasets and pruning rates. (3) We find that HFP is still a reliable choice for large pruning rates.
%(3) We adopt group sparsity and group variance losses during fine-tuning to force neurons to be as different as possible in each layer. 

%%%%%%%%%%%%%%%%%%%%%%%%%%%%%%%%%%%%%%%%%%%%%%%%%%%%%%%%%%

%%%%%%%%%%%%%%%%%%%%%%%%%%%%%%%%%%%%%%%%%%%%%%%%%%%%%%%%%%

\section{Related works}

%%%%%%%%%%%%%%%%%%%%%%%%     GHFP   %%%%%%%%%%%%%%%%%%%%%%%%%%%%%%%%%%
%%%%%%%%%%%%%%%%%%%%%%%%%%%%%%%%%%%%%%%%%%%%%%%%%%%%%%%%%%

Previous works on accelerating CNNs mainly consist of \emph{matrix decomposition}~\cite{Jaderberg2014,Alvarez2017}, \emph{quantization}~\cite{Hubara2016}, \emph{fast convolution}~\cite{Krizhevsky2012,Howard2017,Singh2019}, \emph{knowledge distillation}~\cite{Hinton2015} and \emph{pruning}. 

\emph{Pruning}-based methods focus on removing unimportant connections or neurons to compress and accelerate the network. \emph{Pruning} methods can be divided into two categories: \textbf{weight pruning} and \textbf{filter pruning}. Weight pruning methods compress networks in an unstructured manner by deleting redundant weights, thus inducing sparsity in filters. While weight pruning methods require specialized libraries to obtain real acceleration, filter pruning directly deletes unimportant filters in a structured manner, capable of easily reducing both the model size and the computational overhead~\cite{Zhu2018ToPO, Liu2019RethinkingTV, Frankle2019TheLT, Luo2017}.

During these years, numerous filter pruning methods have been proposed. Prevalent criteria for measuring the importance of a filter or a channel include $\ell_{1}$-norm, $\ell_{2}$-norm, scaling factors, feature redundancy and cross-layer filter comparison~\cite{Li2016,Liu2017,Ayinde2018, Wang2019}. Recently proposed Filter Pruning via Geometric Median (FPGM) method indicates that the widely used  “smaller-norm-less-important” criterion requires significant deviation of filter norms and norms of those filters chosen to be pruned close to zeros, and proposes to prune filters via Geometric Median~\cite{He2019FilterPV}, while Guided Structured Sparsity (GSS) imposes group sparsity and group variance losses to force a portion of neurons  in each layer to stay alive~\cite{Torfi2018AttentionBasedGS}. In effect, GSS pushes filters to have a high variance with a concentration around zero, which is consistent with the requirements of  “smaller-norm-less-important” criterion unintentionally.% Inspired by this coincidence, we adopt a group sparisity and group variance loss during training to obtain a high variance.
 
SFP and ASFP simply set pruned filters to zeros during training, keeping updating them in the following training epochs to maintain a large search space. In contrast, SRFP and ASRFP 
decay pruned filters with a monotonic decreasing parameter in order to alleviate the information loss caused by pruning. Nevertheless, although maintaining a large search space, these soft pruning methods converge much slower than HFP in case of large pruning rates.

Hence, we propose a Gradually Hard Filter Pruning (GHFP) method to smoothly switch from ASP to HFP during training and pruning to strike a balance between performance and convergence speed. We utilize a monotonic increasing parameter to control the ratio of HFP, increasing from 0 to 1, initially acting like ASP to maintain a large search space, gradually increasing the rate of HFP to reduce the search space to ensure convergence.

%%%%%%%%%%%%%%%%%%%%%%%%%%%%%%%%%%%%%%%%
\section{Our method}
\label{sec:prop}

\subsection{Formulation}
\label{sec:idea}

The dimension of the convolutional kernel in the $i$-th layer ${W}_{i}$ can be denoted as $n\times m\times s\times s$, where $1\leq i\leq L$ and $L$ is the total number of convolutional layers in a network. Concretely, $s$ denotes the kernel size. $m$ and $n$ represent the number of input channels and output channels respectively.

The shape of the input tensor ${I}_i$ is $m \times h_{i}\times w_{i} $ and the output tensor ${O}_i \in \mathbb{R}^{n \times h_{i+1}\times w_{i+1}}={W}_{i}\ast{I}_i$ can be written as

\vspace{-3mm}
{\small
	\begin{align}
	\label{eq:1}
	{O}_{i,j}={W}_{i,j}\ast{I}_i ~&~ {for}~1 \leq j \leq n,
	\end{align}
}
\vspace{-3mm}

\noindent where ${O}_{i,j}  \in \mathbb{R}^{h_{i+1}\times w_{i+1}}$ is the $j$-th output channel in the $i$-th layer and ${W}_{i,j} \in \mathbb{R}^{m\times s\times s}$ denotes the $j$-th filter in the $i$-th layer.

Assume the filter pruning rate for the $i$-th layer to be $P_{i}$.
Hence, we will delete $n\times P_{i}$ filters in the $i$-th layer,
and the shape of the pruned output tensor ${O}_{i}$ is ${n\times (1-P_{i})\times h_{i+1} \times w_{i+1}}$.

{\bf Soft Pruning.} Based on the SRFP method and its variant ASRFP, the pruned weights of the $i$-th layer are simply zeroized, which can be represented by

\vspace{-3mm}
{\small
	\begin{align}
	\label{eq:2}
	%\hspace{-3.5mm}
	{\hat{W}}_{i,j}\!=\!{W}_{i,j}\!\odot\!{M}_{i,j}\!+\!
	\alpha\;\! {W}_{i,j}\!\odot\!(1\!-\!{M}_{i,j}) ~&~ {for}~1 \leq j \leq n,
	\end{align}
}
\vspace{-3mm}

\noindent where ${M}_{i,j}$ is a Boolean mask of the filter ${W}_{i,j}$ to represent whether the $j$-th filter in the $i$-th layer is removed.  $\hat{W}_{i,j}$ is the resulted filter. ${M}_{i,j}=0$ if ${W}_{i,j}$ is pruned. Otherwise, ${M}_{i,j}=1$ denotes that the filter ${W}_{i,j}$ is not pruned. $\odot$ denotes the element-wise multiplication. $\alpha$ is a monotonic decreasing parameter to control the decaying speed of pruned filters and to better utilize the trained information of pruned filters. Generally, $\alpha \in [0,1]$. SFP and ASFP are special cases of SRFP and ASRFP respectively when $\alpha=0$.

When $\alpha \in (0,1]$, the trained information of pruned filters is not entirely abandoned. Nevertheless, when $\alpha > 0$, the softly pruned model is not compact, since pruned filters are not zeroized. Hence, SRFP and ASRFP gradually decay $\alpha$ from 1 towards 0 via an exponential decay strategy, given by 

\vspace{-3mm}
{\small
	\begin{align}
	\label{eq:3}
	\alpha_e(t) = \alpha_0(\frac{\alpha_0}{\epsilon})^{-\frac{t}{t_{max}-1}} ~&~ {for}~0 \leq t < t_{max},
	\end{align}
}
\vspace{-3mm}

\noindent where $t_{max}$ is the maximal number of training epochs and $\alpha_0=1$ in SRFP and ASRFP. $\epsilon$ is infinitely close to zero to satisfy that

\vspace{-3mm}
{\small
	\begin{align}
	\label{eq:4}
	\alpha_e(t_{max}-1) = \epsilon \to 0.~&~
	\end{align}
}
\vspace{-3mm}

\noindent When $\alpha_e(t)$ is close to zero enough, SRFP and ASRFP set $\alpha_e(t)=0$ to obtain a really compact model.

{\bf Hard Pruning.} In HFP, the pruning rate, consistent with that of ASRFP, would gradually increase from 0 to the final pruning rate. Therefore, HFP can be seen as a special case of ASRFP when $\alpha=0$, apart from stopping updating pruned filters, represented as

\vspace{-3mm}
{\small
	\begin{align}
	\label{eq:5}
	%\hspace{-3.5mm}
	{\hat{G}}_{i,j}\!=\!{G}_{i,j}\!\odot\!{M}_{i,j} ~&~ {for}~1 \leq j \leq n,
	\end{align}
}
\vspace{-3mm}

\noindent where ${G}_{i,j}$ is the gradient of the filter ${W}_{i,j}$ and $\hat{G}_{i,j}$ is the  masked gradient. Only unpruned filters could update their weights. Thus, for one thing, the representative capacity of the model is gradually reduced. For another, the search space is gradually narrowed to ensure convergence.

\subsection{Gradually Hard Filter Pruning (GHFP)}
\label{sec:network}

Although maintaining
a large search space, soft pruning methods converge
much slower than HFP in case of large pruning rates. 

Therefore, we  propose a Gradually Hard Filter Pruning (GHFP) method to smoothly switch from ASP to HFP during training and pruning to achieve a balance between performance and convergence speed. We adopt a monotonic increasing parameter $\lambda_{h}$ to control the ratio of HFP, increasing from 0 to 1, given by

\vspace{-3mm}
{\small
	\begin{align}
	\label{eq:6}
	%\hspace{-3.5mm}
	\lambda_{h}(t)\!=\!\lambda_{i}\!+\!(\lambda_{f}\!-\!\lambda_{i})\!\left[\!1 \!\!-\!\! (\frac{\!t_{max}\!-\!1\!-\!t}{t_{max-1}})^3\!\right]\! \!\!~&~\! {for}~0\!\leq t\! <\! t_{max},
	\end{align}
}
\vspace{-3mm}

\noindent where $\lambda_i$ and $\lambda_f$ are the initial and final ratio of HFP respectively.  If the pruning rate of the $i$-th layer in the $t$-th epoch is $P_i(t)$, we simply set $P_i(t)\times n$ channels to zeros, among which $\lambda_{h}(t)\times P_i(t)\times n$ channels with minimum $\ell_2$-norms are pruned in a hard manner, while $(1-\lambda_{h}(t))\times P_i(t)\times n$ channels are softly pruned, where $n$ is the number of channels in the $i$-th layer.

\begin{figure}[t]
  \centering
  \includegraphics[width=35mm,clip]{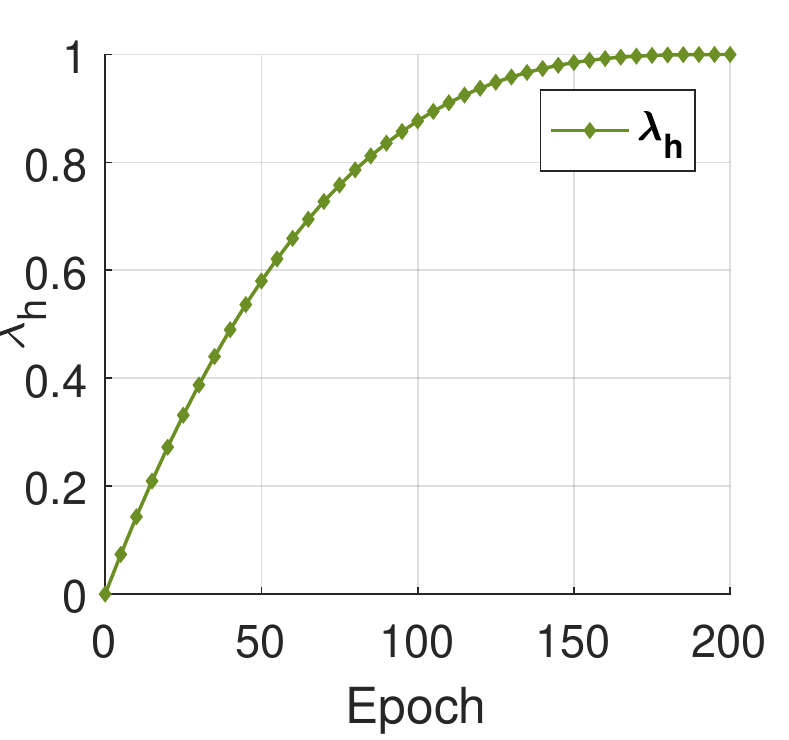}  
  \vspace{-5pt}
  \caption{Gradually increase the proportion of HFP as the training and pruning procedure goes on.}
  \label{fig:lambda_hard}
  \vspace{4pt}
\end{figure}

We simply set $\lambda_{i}=0$, $\lambda_{f}=1$ and  $t_{max}=200$. Thus,  as shown in Figure\,\ref{fig:lambda_hard}, our GHFP initially acts like ASP to maintain a large capacity, gradually increasing the rate of HFP to reduce the search space to ensure convergence. We present our GHFP method in Algorithm~\ref{alg:GHFP}. Especially, if $\alpha_0 = 0$, we called our method ASFP combined with HFP. Otherwise, we called it ASRFP combined with HFP.

\vspace{2mm}
\begin{algorithm}
	\caption{GHFP Algorithm}\label{alg:GHFP}
	\footnotesize
	\SetKwInOut{Input}{inputs}\SetKwInOut{Output}{output}
	
	\Input{training set: ${X}$, final pruning rate: $P_{i}$, initial decay rate: $\alpha_0$, the model with parameters ${W} = \{{W}_{i}, 0\leq i \leq L\}$. }
	\Output{The pruned model with parameters ${W} ^{*}={W} ^{t_{max}}$}
	%\vspace{0.1cm}
	Initialize the model parameter ${W}^0$\\
	Initialize HFP rate $\lambda_{h}(0)=0$\\
	Initialize Pruning rate $P_{i}(0)=0$\\
	\For{$t=0$, ..., $ t_{max}-1$}{
		Decrease weight decay rate $\alpha$  with  Eq.\eqref{eq:3}\\
		Increase HFP rate $\lambda_{h}(t)$ with  Eq.\eqref{eq:6}\\
			Increase pruning rate $P_{i}(t)$ based on ASFP\\
		Train model parameters $\hat{W}^{t+1}$ based on data set ${X}$ and  ${W}^{t}$\\
		\For{$i=1$, ..., $L$}{
			Compute the $\ell_2$-norm of each filter $\|\hat{W}_{i,j}^{t+1}\|_2, 1 \leq j \leq n$ \\
			Select $n\times P_i$ filters with minimal $\ell_2$-norm values\\
			$\lambda_{h}(t)\times P_i(t)\times n$  filters with minimum $\ell_2$-norms are pruned hard\\
			 $(1-\lambda_{h}(t))\times P_i(t)\times n$ remaining chosen filters are softly pruned using $\alpha$\\
			
		}
		Get the pruned model parameters ${W}^{t+1}$ based on $\hat{W}^{t+1}$
		
	}
	Get the pruned model with final parameters ${W} ^{*}={W} ^{t_{max}}$  \\
\end{algorithm}
%\vspace{-2mm}

%%%%%%%%%%%%%%%%%%%%%%%%%%%%%%%%%%%%%%%%%%%%%%%%%%%%%%%%%%
%%%%%%%%%%%%%%%%%%%%%%%%%%%%%%%%%%%%%%%%%%%%%%%%%%%%%%%%%%

%{\bf Loss Function.} 
%We adopt group sparsity and group variance losses during fine-tuning to push neurons to be different in each layer.
%calculated as
%\begin{align}
%\mathcal{L}_{gs}(l) &= \sum_{i=1}^{C}\sqrt{\sum_{j=1}^{|w_i|}(w_{ij})^2},\\
%\mathcal{L}_{gv}(l) &= 
%\frac{1}{C}\sum_{i=1}^{C}
%\left(
%\sqrt{\sum_{j=1}^{|w_i|}(w_{ij})^2}
%- \frac{1}{C}\sum_{k=1}^{C}\sqrt{\sum_{j=1}^{|w_k|}(w_{kj})^2}
%\right)^2,
%\end{align}
%where  $C$ is the number of channels in the $l$-th layer and and $|w_{ij}|$ is the number of weights in $w_{i}$.

%%%%%%%%%%%%%%%%%%%%%%%%%%%%%%%%%%%%%%%%
\section{Experiments}
\label{sec:exp}

\subsection{Setup}
\label{sec:exp_varif}

Our method is evaluated on CIFAR-10/100~\cite{Krizhevsky2009}. CIFAR-10 and CIFAR-100 include 50,000 training images and 10,000 test images of size $32 \times 32$ pixels, divided into 10 and 100 classes respectively. We mainly prune the challenging ResNet, following the experimental setup in CP~\cite{He2017}, ASFP~\cite{He2019}. We also prune the VGG-16~\cite{Simonyan2015} following the setup in~\cite{Li2016,Liu2017,He2019}.

On CIFAR-10/100, we follow the parameter scheme and the training configuration in~\cite{He201602, He2019}.

Our GHFP is adopted after each training epoch. Models are trained from scratch by default. Results starting from pre-trained models are also provided, where the learning rate is one tenth of that of models trained from scratch. Experiments are repeated four times. %, reported by the ``mean $\pm$ std''.
Then we compare our results with other state-of-the-art methods, e.g., SFP~\cite{He2018}, SRFP~\cite{Cai2020}, ASRFP~\cite{Cai2020},  ASFP~\cite{He2019}, MIL~\cite{Dong2017}, PFEC~\cite{Li2016}, CP~\cite{He2017}, FPGM~\cite{He2019FilterPV},	VCNNP~\cite{Zhao2019VariationalCN}.

%\vspace{0mm}

\subsection{Results on CIFAR-10}
\label{sec:exp_varif}

\textbf{Settings}. On CIFAR-10, our GHFP is evaluated on VGG-16 and ResNet-20/56/110. %By default, we use exponential decay with $\epsilon=1e-5$ and $\alpha_0=1$.

\textbf{Results}. We summarize the results on CIFAR-10 
in Table~\ref{table:cifar10_20_110_acc}, Table~\ref{table:cifar10_56_acc} and Table~\ref{table:cifar10_vgg_acc}, where "Our1" and "Our2" denote ASFP combined with HFP and ASRFP combined with HFP respectively. "PR" means the pruning rate. Models are trained from scratch by default. Both GHFP methods achieve competitive performance on %both
CIFAR-10 compared with other channel pruning methods across networks of various depths and pruning rates. Particularly, ASFP combined with HFP outperforms other methods in most cases, while the performance of ASRFP combined with HFP fluctuates obviously.

%%---------------   CIFAR-10  resnet20-110-------------%%
%% CIFAR-10
\begin{table}[th]
	\setlength{\tabcolsep}{1mm}
	\caption{Pruning results of ResNet-20/110 on CIFAR-10.
	} 	
	%\vspace{-0mm}
	\label{table:cifar10_20_110_acc} 
	\small  
	\footnotesize
	\centering  	
	\begin{tabular}{c | c c c c}  		
		\toprule %\hline 		
		 Depth & Method  &Baseline(\%)  &Accu.(\%)  & FLOPs(PR\%)    
		\\ \midrule %\hline 

			\multirow{10}{*}{20} &	 MIL~\cite{Dong2017} &
		 91.53                   & 91.43   &3.20E7(20.3) \\
		%&\vspace{-4mm} \ldotfill{4pt}{1pt} & \ldotfill{14pt}{11pt} & \ldotfill{14pt}{11pt}&
		%\ldotfill{14pt}{11pt}& \ldotfill{14pt}{11pt} & \ldotfill{14pt}{11pt}	 \\ 

		 &SFP(20\%)  &{92.32}   & 91.40   &2.87E7(29.3)\\
		 &SRFP(20\%) &{92.32}   & 91.47   &  2.87E7(29.3)\\\              
		 &ASFP(20\%) &{92.89}  & 91.62  &  2.87E7(29.3)\\              
		 &ASRFP(20\%)  &{92.89}   & 91.67   &  2.87E7(29.3)\\

		 &Our1(20\%)   &{92.89}  & 92.14  &  2.87E7(29.3)\\      
		&\vspace{-4mm} \ldotfill{4pt}{1pt} & \ldotfill{14pt}{11pt}&
		\ldotfill{14pt}{11pt}& \ldotfill{14pt}{11pt}	 \\            
		 &Our2(20\%)  &{92.89}   & \textbf{92.41}  &  2.87E7(29.3)\\

		%&\vspace{0.2mm} \ldotfill{4pt}{1pt}  & \ldotfill{14pt}{11pt} & \ldotfill{14pt}{11pt}&
		%\ldotfill{14pt}{11pt}& \ldotfill{14pt}{11pt} & \ldotfill{14pt}{11pt}	 \\ 
		
				%& SFP(20\%) &$\times$  &{92.32}   & 91.40   &2.87E7 &29.3\\
		%& SRFP(40\%) &$\times$  &{92.32}   & 90.25   & 2.43E7 &54.0 \\             
		 &ASFP(40\%) &{92.89}  & 90.57  &  2.43E7(54.0) \\              
		 &ASRFP(40\%) &{92.89}   & 90.65   & 2.43E7(54.0) \\  
		%&\vspace{-4mm} \ldotfill{4pt}{1pt} & \ldotfill{14pt}{11pt}  & \ldotfill{14pt}{11pt}&
		%\ldotfill{14pt}{11pt}& \ldotfill{14pt}{11pt} 	 \\              
		
		&Our1(40\%)  &{92.89}  & 90.82  & 2.43E7(54.0) \\    			\vspace{-0mm} %&\ldotfill{4pt}{1pt} & \ldotfill{14pt}{11pt}&
		%\ldotfill{14pt}{11pt}& \ldotfill{14pt}{11pt}	 \\        
		&Our2(40\%)&{92.89}   & \textbf{91.28}  & 2.43E7(54.0) \\  
		
			\midrule %\hline  
       %%%%%%%%%%% cifar10 resnet 110 %%%%%%%%%%%%%%%
     
     		\multirow{8}{*}{110} &PFEC~\cite{Li2016} &93.53     & 92.94      & 1.55E8(38.6) 	 \\       
	     %&\vspace{-4mm} \ldotfill{4pt}{1pt} & \ldotfill{14pt}{11pt}  & \ldotfill{14pt}{11pt}&
	     %\ldotfill{14pt}{11pt}& \ldotfill{14pt}{11pt} & \ldotfill{14pt}{11pt}	 \\   
	     &MIL~\cite{Dong2017} & 93.63         & 93.44&	-(34.2) 	 \\           		
	     &SFP(20\%)   & {94.33}	&{93.86}  &  1.82E8(28.2) \\
	     
	     &SRFP(20\%) & {94.33}	&{93.61} &    1.82E8(28.2) \\
	     &ASFP(20\%)  & {94.76}	&{94.71} &    1.82E8(28.2) \\ 
	     &ASRFP(20\%) & {94.76}	&{95.10} &    1.82E8(28.2) \\
	     &Our1(20\%)  & {94.76}	&\textbf{95.16} &    1.82E8(28.2) \\
	     %&\vspace{-4mm} \ldotfill{4pt}{1pt} & \ldotfill{14pt}{11pt} &
	     %\ldotfill{14pt}{11pt}& \ldotfill{14pt}{11pt}& \ldotfill{14pt}{11pt} & \ldotfill{14pt}{11pt}	 \\ 
	     &Our2(20\%)   & {94.76}	&{94.59} &   1.82E8(28.2) \\

   		 \bottomrule %\hline  	
	\end{tabular}
	%\vspace{-2mm}  	
\end{table}

%%---------------   CIFAR-10  resnet56--------------%%
%% CIFAR-10
\begin{table}[th]
	\setlength{\tabcolsep}{1mm}
	\caption{Pruning results of ResNet-56 on CIFAR-10. ${\star}$ denotes pruning from pre-trained models.
	} 	
	%\vspace{-0mm}
	\label{table:cifar10_56_acc} 
	\small  
	\footnotesize
	\centering  	
	\begin{tabular}{c c c c}  		
		\toprule %\hline 		
		Method  &Baseline(\%)  &Accu.(\%)  & FLOPs(PR\%)    
		\\ \midrule %\hline 

		%%%%%%%%%%%%%%    cifar-10 resnet56   %%%%%%%%%%%%%%%
		PFEC~\cite{Li2016} & 93.04                   & 91.31   &9.09E7(27.6) \\
		%&\vspace{-4mm} \ldotfill{4pt}{1pt} & \ldotfill{14pt}{11pt} & \ldotfill{14pt}{11pt}&
		%\ldotfill{14pt}{11pt}& \ldotfill{14pt}{11pt} & \ldotfill{14pt}{11pt}	 \\ 
		
		CP~\cite{He2017} & 92.80                      & 90.90 &	-(50.0)\\   
		
		SFP(20\%)   &{93.66}   & 93.26&8.98E7(28.4)\\  
		SRFP(20\%) &{93.66}   & {93.33}  &8.98E7(28.4)\\             
		ASFP(20\%)  &{94.85}  & 93.97  &8.98E7(28.4)\\

		ASRFP(20\%) &{94.85}  & 93.77     &8.98E7(28.4)\\     
		Our1(20\%)  &{94.85}  & \textbf{94.17}    &8.98E7(28.4)\\     
		\vspace{-4mm} \ldotfill{14pt}{11pt} &  \ldotfill{14pt}{11pt}&
		\ldotfill{14pt}{11pt}& \ldotfill{14pt}{11pt}	 \\    
		Our2(20\%)   &{94.85}  & 94.10    &8.98E7(28.4)\\     
		
		PFEC${}^{\star}$~\cite{Li2016} & 93.04                   & 93.06   &9.09E7(27.6) \\
		SFP(20\%)${}^{\star}$   &{93.66}   & 93.25&8.98E7(28.4)\\  
		ASFP(20\%)${}^{\star}$  &{94.85}  & 94.92  &8.98E7(28.4)\\     
		%& SRFP(20\%) &\checkmark  &{93.34}  & 93.17  &8.98E7(28.4)\\     
		ASRFP(20\%)${}^{\star}$   &{94.85}  & 94.98    &8.98E7(28.4)\\     
		
		Our1(20\%)${}^{\star}$ &{94.85}  & \textbf{95.07}   &8.98E7(28.4)\\     
		
		\vspace{-4mm} \ldotfill{14pt}{11pt} & \ldotfill{14pt}{11pt}&
		\ldotfill{14pt}{11pt}& \ldotfill{14pt}{11pt}	 \\ 
		
		Our2(20\%)${}^{\star}$ &{94.85}  & 94.92    &8.98E7(28.4)\\

		%& SFP(40\%) &$\times$  &{93.66}   & 92.06 &5.94E7 &52.6\\
		%& SRFP(40\%) &$\times$  &{93.66}   & {92.67}  &5.94E7 &52.6\\                
		ASFP(40\%) &{94.85}  & 93.64   & 5.94E7(52.6) \\     
		
		ASRFP(40\%)  &{94.85}  & 93.68    & 5.94E7(52.6) \\      
		Our1(40\%) &{94.85}   & \textbf{93.84}& 5.94E7(52.6) \\ 
		
		\vspace{-4mm} \ldotfill{14pt}{11pt} & \ldotfill{14pt}{11pt}&
		\ldotfill{14pt}{11pt}& \ldotfill{14pt}{11pt}	 \\ 
		        
		Our2(40\%)  &{94.85}  & {93.75}   & 5.94E7(52.6) \\

		ASFP(60\%)${}^{\star}$ &{94.85}  & 89.72   & 3.43E7(72.6) \\     
		
		ASRFP(60\%)${}^{\star}$  &{94.85}  & 90.54    &3.43E7(72.6) \\     
		Our1(60\%)${}^{\star}$ &{94.85}   & 92.54 &3.43E7(72.6) \\     
		Our2(60\%)${}^{\star}$  &{94.85}  & \textbf{93.00}   &3.43E7(72.6) \\

		\bottomrule %\hline  	
	\end{tabular}
	\vspace{-5.mm}  	
\end{table}

%%---------------   CIFAR-10  resnet110--------------%%
%% CIFAR-10
%\begin{table}[th]
%	\setlength{\tabcolsep}{1mm}
%	\caption{Pruning results of ResNet-110 on CIFAR-10.
%	} 	
%	%\vspace{-0mm}
%	\label{table:cifar10_110_acc} 
%	\small  	
%	\centering  	
%	\begin{tabular}{c c c c c}  		
%		\toprule %\hline 		
%		Method  &Baseline(\%)  &Accu.(\%)  & FLOPs (PR\%)    
%		\\ \midrule %\hline 
%		
%		
%		PFEC~\cite{Li2016} &93.53     & 92.94      & 1.55E8(38.6) 	 \\       
%		%&\vspace{-4mm} \ldotfill{4pt}{1pt} & \ldotfill{14pt}{11pt}  & \ldotfill{14pt}{11pt}&
%		%\ldotfill{14pt}{11pt}& \ldotfill{14pt}{11pt} & \ldotfill{14pt}{11pt}	 \\   
%		MIL~\cite{Dong2017} & 93.63         & 93.44&	-(34.2) 	 \\           		
%		SFP(20\%)   & {94.33}	&{93.86}  &  1.82E8(28.2) \\
%		
%		SRFP(20\%) & {94.33}	&{93.61} &    1.82E8(28.2) \\
%		ASFP(20\%)  & {94.76}	&{94.71} &    1.82E8(28.2) \\ 
%		ASRFP(20\%) & {94.76}	&{95.10} &    1.82E8(28.2) \\
%		Our1(20\%)  & {94.76}	&\textbf{95.16} &    1.82E8(28.2) \\
%		%&\vspace{-4mm} \ldotfill{4pt}{1pt} & \ldotfill{14pt}{11pt} &
%		%\ldotfill{14pt}{11pt}& \ldotfill{14pt}{11pt}& \ldotfill{14pt}{11pt} & \ldotfill{14pt}{11pt}	 \\ 
%		Our2(20\%)   & {94.76}	&{94.59} &   1.82E8(28.2) \\
%		
%		\bottomrule %\hline  	
%	\end{tabular}
%	%\vspace{-2mm}  	
%\end{table}  

%%---------------   CIFAR-10  vgg-16--------------%%
%% CIFAR-10
\begin{table}[htb]
	\setlength{\tabcolsep}{1mm}
	\caption{Pruning results of VGG-16 on CIFAR-10. ${\star}$ denotes pruning from pre-trained models.
	} 	
	\vspace{-0mm}
	\label{table:cifar10_vgg_acc} 
	%\small 
	\footnotesize 	
	\centering  	
	\begin{tabular}{c c c c c}  		
		\toprule %\hline 		
		Method  &Baseline(\%)  &Accu.(\%)  & FLOPs (PR\%)    
		\\ \midrule %\hline 

		PFEC~\cite{Li2016} &93.58     & 93.31      & 2.04E8(34.2) \\
		FPGM~\cite{He2019FilterPV} &93.58     & 93.23      & 1.99E8(35.9) \\      
		VCNNP~\cite{Zhao2019VariationalCN} &93.25     & 93.18      & 1.90E8(39.1) \\
		%\ldotfill{14pt}{11pt}& \ldotfill{14pt}{11pt} & \ldotfill{14pt}{11pt}	 \\   
		%MIL~\cite{Dong2017} & 93.63         & 93.44&	-(34.2) 	 \\           		
		SFP(20\%)  & {93.62}	&{93.11}  &  2.04E8(34.2) \\
		
		%SRFP(20\%) & {73.69}	&{93.61} &    1.82E8(28.2) \\
		%ASFP(50\%)  & {93.62}	&{93.71} & 2.04E8(34.2) \\    
		%ASRFP(50\%) & {93.62}	&{93.40}& 2.04E8(34.2) \\   
		Our1(20\%)  &  {93.62}	&\textbf{93.34} & 2.04E8(34.2) \\   
	
		\vspace{-4mm} \ldotfill{4pt}{1pt} & \ldotfill{14pt}{11pt}&
		\ldotfill{14pt}{11pt}& \ldotfill{14pt}{11pt}	 \\           
		
		Our2(20\%)   & {93.62}	&{93.29}& 2.04E8(34.2) \\   
		
		PFEC${}^{\star}$~\cite{Li2016} &93.58     & 93.28      & 2.04E8(34.2) \\
		SFP(20\%)${}^{\star}$   &{93.62} &{93.69}& 2.04E8(34.2) \\ 
		ASFP(20\%)${}^{\star}$   &{93.62} &{93.75}& 2.04E8(34.2) \\ 
		ASRFP(20\%)${}^{\star}$   &{93.62} &{93.76}& 2.04E8(34.2) \\     
		
		Our1(20\%)${}^{\star}$   &{93.62} &\textbf{93.96}& 2.04E8(34.2) \\ 
		Our2(20\%)${}^{\star}$   &{93.62} &{93.76}& 2.04E8(34.2) \\ 
		
		\bottomrule %\hline  	
	\end{tabular}
	%\vspace{-2mm}  	
\end{table}

\subsection{Results on CIFAR-100}
\label{sec:exp_varif}

\textbf{Settings}. On CIFAR-100, our GHFP is evaluated on ResNet-20/56/110. %By default, we use exponential decay with $\epsilon=1e-5$ and $\alpha_0=1$.

\textbf{Results}. We conclude the results on CIFAR-100
in Table~\ref{table:cifar100_resnet_acc}, where "Our1" and "Our2" denote ASFP combined with HFP and ASRFP combined with HFP respectively. Both GHFP methods still achieve competitive performance. Especially, ASRFP combined with HFP obviously outperforms other methods on CIFAR-100.

%%---------------   CIFAR-100  resnet20/56/110--------------%%
%% CIFAR-10
\begin{table}[th]
	\setlength{\tabcolsep}{1mm}
	\caption{Pruning results of ResNet-20/56/110 on CIFAR-100.
	} 	
	%\vspace{-0mm}
	\label{table:cifar100_resnet_acc} 
	\small 
	\footnotesize	
	\centering  	
	\begin{tabular}{c | c c c c}  		
		\toprule %\hline 		
		Depth & Method  &Baseline(\%)  &Accu.(\%)  & FLOPs(PR\%)    
		\\ \midrule %\hline 

		\multirow{6}{*}{20} &SFP(20\%) &{68.11}   & 66.23   &  2.87E7(29.3)\\\ 
		&SRFP(20\%) &{68.11}   & 66.67   &  2.87E7(29.3)\\\              
		&ASFP(20\%) &{68.92}  & 66.95  &  2.87E7(29.3)\\              
		&ASRFP(20\%)  &{68.92}   & 66.48   &  2.87E7(29.3)\\

		&Our1(20\%)   &{68.92}  & 66.85  &  2.87E7(29.3)\\      
		%\vspace{-4mm} \ldotfill{4pt}{1pt} & \ldotfill{14pt}{11pt}&
		%\ldotfill{14pt}{11pt}& \ldotfill{14pt}{11pt}	 \\            
		&Our2(20\%)  &{68.92}   & \textbf{67.62}  &  2.87E7(29.3)\\

		\midrule %\hline 

		%&\vspace{0.2mm} \ldotfill{4pt}{1pt}  & \ldotfill{14pt}{11pt} & \ldotfill{14pt}{11pt}&
		%\ldotfill{14pt}{11pt}& \ldotfill{14pt}{11pt} & \ldotfill{14pt}{11pt}	 \\ 
		
		%& SFP(20\%) &$\times$  &{92.32}   & 91.40   &2.87E7 &29.3\\
		%& SRFP(40\%) &$\times$  &{92.32}   & 90.25   & 2.43E7 &54.0 \\    
		
			\multirow{6}{*}{56}&SFP(20\%) &{72.00}   & 70.49   &  8.98E7(28.4)\\\ 
		&SRFP(20\%) &{72.00}   & 70.39   &   8.98E7(28.4)\\\               
		&ASFP(20\%) &{72.92}  & 71.01  &   8.98E7(28.4)\\\       
		&ASRFP(20\%)  &{72.92}   &  71.24  &   8.98E7(28.4)\\\

		&Our1(20\%)   &{72.92}  & 71.46  &   8.98E7(28.4)\\\    
		%\vspace{-4mm} \ldotfill{4pt}{1pt} & \ldotfill{14pt}{11pt}&
		%\ldotfill{14pt}{11pt}& \ldotfill{14pt}{11pt}	 \\            
		&Our2(20\%)  &{72.92}   & \textbf{71.93}  &  8.98E7(28.4)\\

		\midrule %\hline  
		%%%%%%%%%%% cifar100 resnet 110 %%%%%%%%%%%%%%%
		
		\multirow{6}{*}{110} &	SFP(20\%) &{73.85}   & 72.66   &  1.82E8(28.2)\\
		&SRFP(20\%)  &{73.85}   & 73.00   &  1.82E8(28.2)\\            
		&ASFP(20\%)  &{74.39}   & 72.91   &  1.82E8(28.2)\\          
		&ASRFP(20\%)  &{74.39}   & 73.02   &  1.82E8(28.2)\\

		&Our1(20\%)    &{74.39}   & 72.95   &  1.82E8(28.2)\\    
		%\vspace{-4mm} \ldotfill{4pt}{1pt} & \ldotfill{14pt}{11pt}&
		%\ldotfill{14pt}{11pt}& \ldotfill{14pt}{11pt}	 \\            
		&Our2(20\%)   &{74.39}   & \textbf{73.29}   &  1.82E8(28.2)\\

		\bottomrule %\hline  	
	\end{tabular}
	%\vspace{-2mm}  	
\end{table}

%%---------------   CIFAR-100  resnet20--------------%%
%% CIFAR-10
%\begin{table}[htb]
%	\setlength{\tabcolsep}{1mm}
%	\caption{Pruning results of ResNet-20 on CIFAR-100.
%	} 	
%	%\vspace{-0mm}
%	\label{table:cifar100_20_acc} 
%	\small  	
%	\centering  	
%	\begin{tabular}{c c c c}  		
%		\toprule %\hline 		
%		Method  &Baseline(\%)  &Accu.(\%)  & FLOPs(PR\%)    
%		\\ \midrule %\hline 
%		
%		
%		SFP(20\%) &{68.11}   & 66.23   &  2.87E7(29.3)\\\ 
%		SRFP(20\%) &{68.11}   & 66.67   &  2.87E7(29.3)\\\              
%		ASFP(20\%) &{68.92}  & 66.95  &  2.87E7(29.3)\\              
%		ASRFP(20\%)  &{68.92}   & 66.48   &  2.87E7(29.3)\\  
%		
%		
%		Our1(20\%)   &{68.92}  & 66.85  &  2.87E7(29.3)\\      
%		%\vspace{-4mm} \ldotfill{4pt}{1pt} & \ldotfill{14pt}{11pt}&
%		%\ldotfill{14pt}{11pt}& \ldotfill{14pt}{11pt}	 \\            
%		Our2(20\%)  &{68.92}   & \textbf{67.62}  &  2.87E7(29.3)\\
%			
%		
%		\bottomrule %\hline  	
%	\end{tabular}
%	%\vspace{-2mm}  	
%\end{table}  

\begin{figure}[H]
	\center
	\vspace{-0mm}
	\subfigure[ASFP combined with HFP]{
		\label{fig:soft_and_to_hard_our1}
		\includegraphics[width=0.52\linewidth]{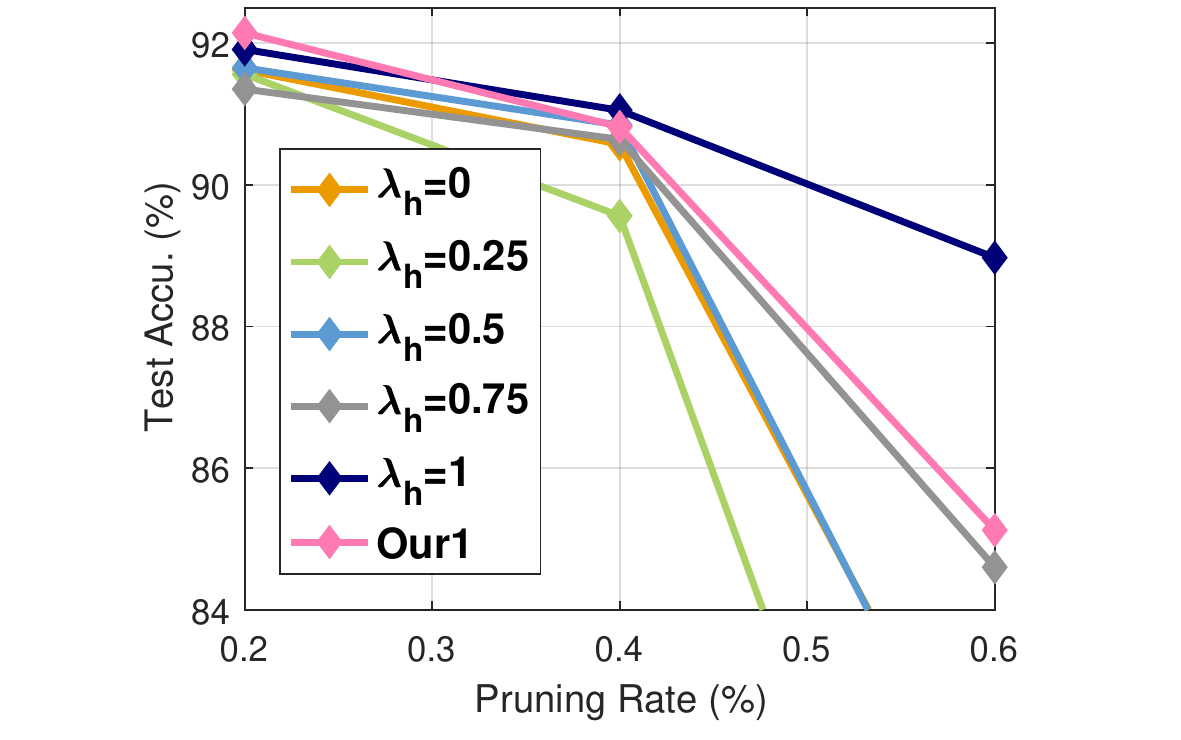}
	}
	\hspace{-9mm}
	\subfigure[ASRFP combined with HFP]{
		\label{fig:soft_and_to_hard_our2}
		\includegraphics[width=0.52\linewidth]{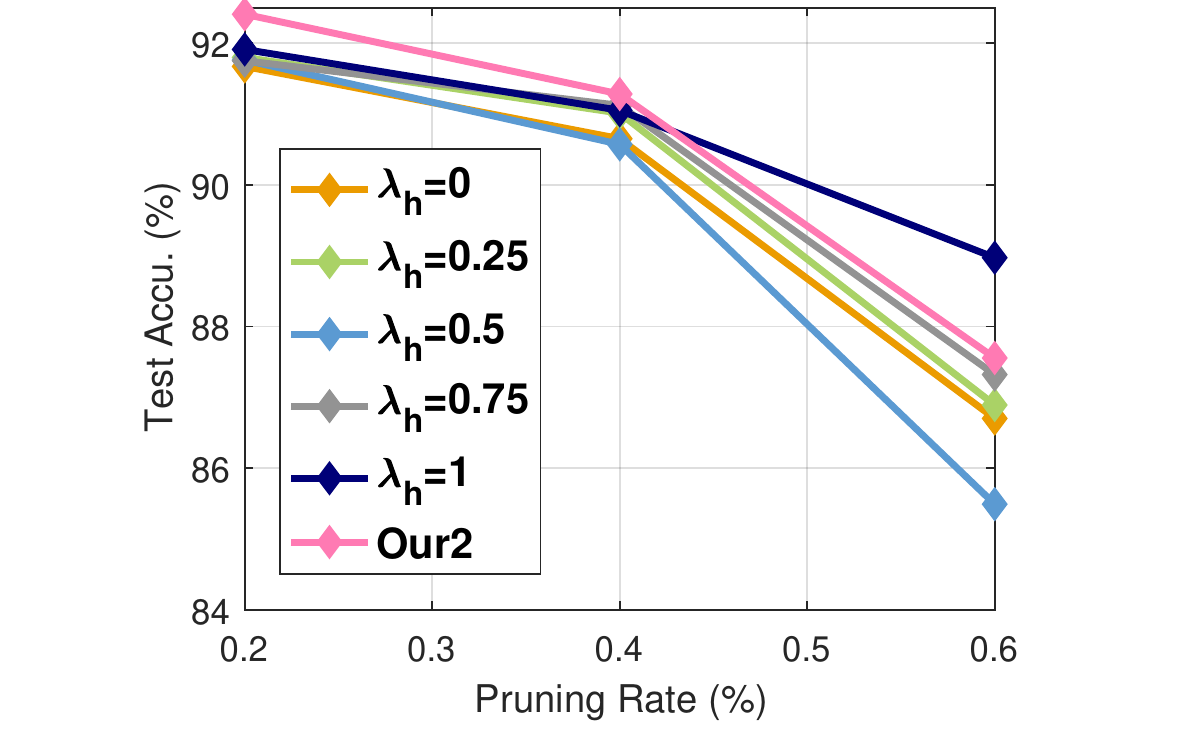}
	}
	\vspace{-2mm}
	\caption{
		Comparison of different $\lambda_h$ for ResNet-20 on CIFAR-10 among "Soft and Hard", ASFP combined with HFP(denoted by "Our1") and ASRFP combined with HFP(denoted by "Our2"). $\lambda_{h}=1$ exactly means HFP while $\lambda_{h}=0$ exactly means ASP.
	}
	\vspace{-3mm}
	\label{fig:soft_and_to_hard}
\end{figure}

\begin{figure}[htb]
	\center
	\vspace{-2mm}
	\subfigure[ResNet-20]{
		\label{fig:soft_to_hard_resnet20}
		\includegraphics[width=0.6\linewidth]{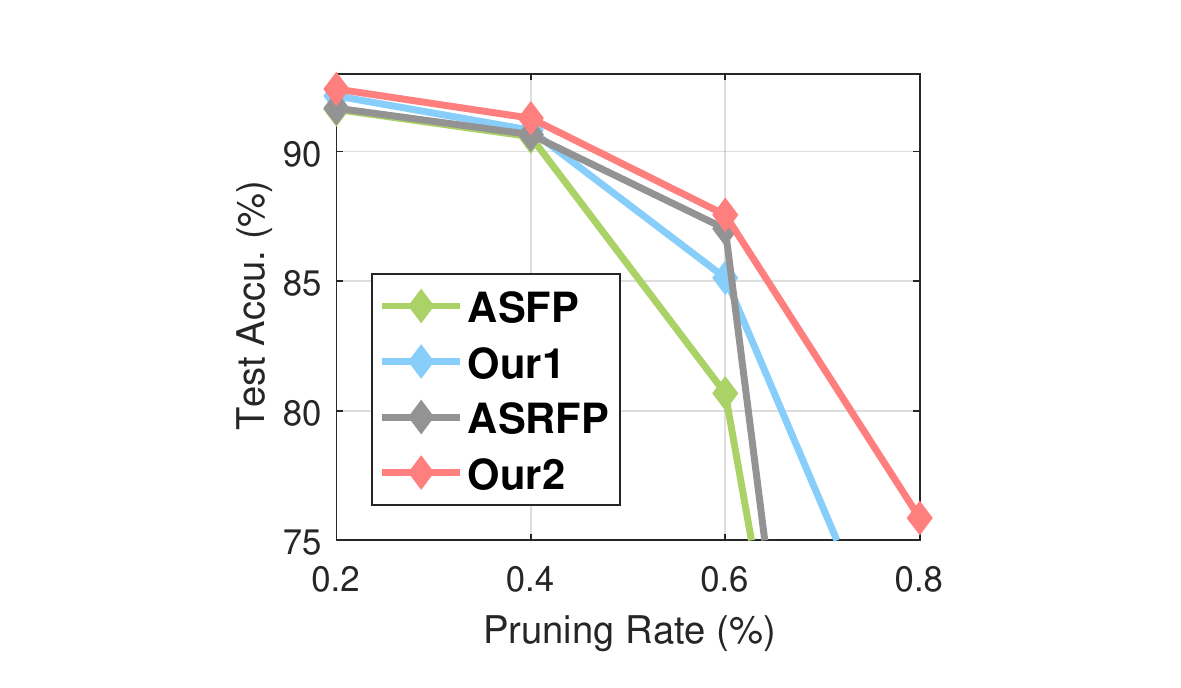}
	}
	\hspace{-22mm}
	\subfigure[ResNet-56]{
		\label{fig:soft_to_hard_resnet56}
		\includegraphics[width=0.6\linewidth]{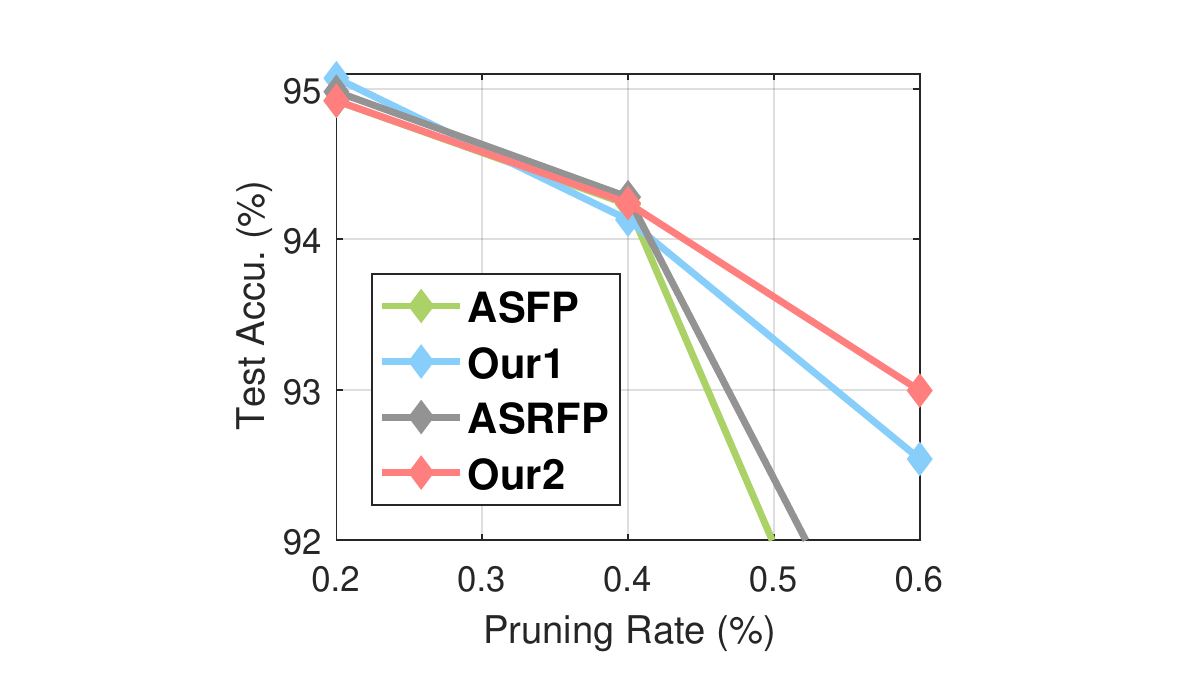}
	}
	\vspace{-2mm}
	\caption{
		Comparison of test accuracies of different pruning rates for ResNet-20/56 on CIFAR-10 among ASFP, ASRFP, ASFP combined with HFP(denoted by "Our1") and ASRFP combined with HFP(denoted by "Our2").
	}
	\vspace{-2mm}
	\label{fig:soft_to_hard_resnet}
\end{figure}

\subsection{Ablation study}
\label{sec:exp_varif}

Extensive ablation experiments of are conducted to analyze our GHFP.

\textbf{Soft and Hard v.s. Soft to Hard.}
We compare results of using a constant $\lambda_{h}$ during training, named "Soft and Hard" and our GHFP, called "Soft to Hard", because a constant $\lambda_{h} \in (0,1)$ makes some filters hard pruned and some filters pruned softly. As shown in Figure~\ref{fig:soft_and_to_hard_our1} and Figure~\ref{fig:soft_and_to_hard_our2}, GHFP-based methods, ASFP combined with HFP and ASRFP combined with HFP outperform "Soft and Hard" methods in terms of performance and stability. Besides, even though our GHFP outperforms soft filter pruning methods, HFP is still a remarkable choice in case of large pruning rates. Our GHFP is more likely to achieve better performance in case of small pruning rates. 

\textbf{Varying pruning rates.}
To further investigate the efficacy of our GHFP method, we present test accuracies of different pruning rates for ResNet-20/56 in Figure~\ref{fig:soft_to_hard_resnet20} and Figure~\ref{fig:soft_to_hard_resnet56}, where ResNet-20 is trained from scratch and ResNet-56 is trained from a pre-trained model. As the pruning rate increases, the test accuracies of GHFP-based methods decline much steadier than those of ASFP and ASRFP. In both cases, our GHFP-based methods outperform ASFP and ASRFP in terms of stability and test accuracy across diverse pruning rates.

\section{Conclusions}
\label{sec:cncl}

We analyze the characteristics of soft pruning methods and HFP, noting that although pursuing better performance, soft pruning methods may be confronted with severe test accuracy drops in case of large pruning rates because they don't reduce their search space to ensure convergence. Hence, we propose a method named GHFP to smoothly switch from ASP to HFP during training and pruning to achieve a balance between performance and convergence speed. Our GHFP performs well across various networks, datasets and pruning rates. Besides, HFP is a reliable choice for large pruning rates.

\bibliographystyle{IEEEbib}
\bibliography{ghfp}

\end{document}